\documentclass[letterpaper, 10 pt, conference]{ieeeconf}  

\IEEEoverridecommandlockouts                              

\usepackage{graphics} 
\usepackage{epsfig} 
\usepackage{mathptmx} 
\usepackage{times} 
\usepackage{amsmath} 
\usepackage{amssymb}  
\usepackage{algorithm}  
\usepackage{algorithmic}
\usepackage{colortbl} 
\usepackage{booktabs}
 \usepackage{multirow} 
\usepackage{xcolor}
\definecolor{grey}{rgb}{128,128,128}
\title{\LARGE \bf
Embodied Domain Adaptation for Object Detection
}

\author{Xiangyu Shi, Yanyuan Qiao, Lingqiao Liu, and Feras Dayoub
\thanks{The authors are with the School of Computer Science and the Australian
Institute for Machine Learning at the University of Adelaide, Adelaide,
Australia (xiangyu.shi@adelaide.edu.au). }
\thanks{The authors thank Zheng Yu and Jiajun Deng for insightful discussions, and Wenqi Lyu for structural suggestions.}}

\begin{document}

\maketitle
\thispagestyle{empty}
\pagestyle{empty}

\begin{abstract}

Mobile robots rely on object detectors for perception and object localization in indoor environments. However, standard closed-set methods struggle to handle the diverse objects and dynamic conditions encountered in real homes and labs. Open-vocabulary object detection (OVOD), driven by Vision Language Models (VLMs), extends beyond fixed labels but still struggles with domain shifts in indoor environments. We introduce a Source-Free Domain Adaptation (SFDA) approach that adapts a pre-trained model without accessing source data. We refine pseudo labels via temporal clustering, employ multi-scale threshold fusion, and apply a Mean Teacher framework with contrastive learning. Our Embodied Domain Adaptation for Object Detection (EDAOD) benchmark evaluates adaptation under sequential changes in lighting, layout, and object diversity. Our experiments show significant gains in zero-shot detection performance and flexible adaptation to dynamic indoor conditions.

\end{abstract}


\section{INTRODUCTION}


Robust object detection is pivotal for mobile robots performing tasks like semantic mapping, navigation, and object interaction in indoor environments. While standard object detectors (e.g., those trained on COCO~\cite{coco}) excel at detecting fixed, closed-set categories, their applicability is limited in real-world homes or labs, which often contain a vast array of objects beyond any predefined label set.  Open-vocabulary object Detection (OVOD) addresses this challenge by leveraging Vision Language Models (VLMs)~\cite{detic,yoloworld,gdino}. Although these models demonstrate robust performance on large-scale benchmarks, they often experience notable accuracy drops under the dynamic lighting conditions and complex layouts characteristic of indoor robotics scenarios.

Such performance drops frequently stem from a domain mismatch between the environment used for training and the target indoor setting. Domain Adaptation~\cite{chen2018domain,saito2019strongweak,sindagi2020prior} offers a cost-effective alternative to retraining from scratch by adjusting a model trained in a source domain to perform better in a target domain. Although many approaches rely on data from both source and target domains~\cite{udaa}, practical constraints such as data privacy policies or proprietary restrictions may prevent access to the source dataset. When adaptation must proceed without access to original training data, the problem becomes Source-Free Domain Adaptation (SFDA)~\cite{vs2023towards,tarvainen2017mean,yangyang}.

The existing SFDA in Object Detection (SFOD) approach typically considers using shared categories across the source domain and target domain. The datasets used for SFOD are mainly related to autonomous driving, such as Cityscapes~\cite{cityscape}, Sim10k~\cite{sim10k}, and SHIFT~\cite{shift2022} or traditional Object Detection datasets, such as COCO~\cite{coco}, VOC2017~\cite{VOC}. To the best of our knowledge, there is currently no benchmark dedicated to SFOD in indoor, non-stationary environments.
\begin{figure}[t]
    \centering
    \includegraphics[width=\linewidth, height=0.55\linewidth]{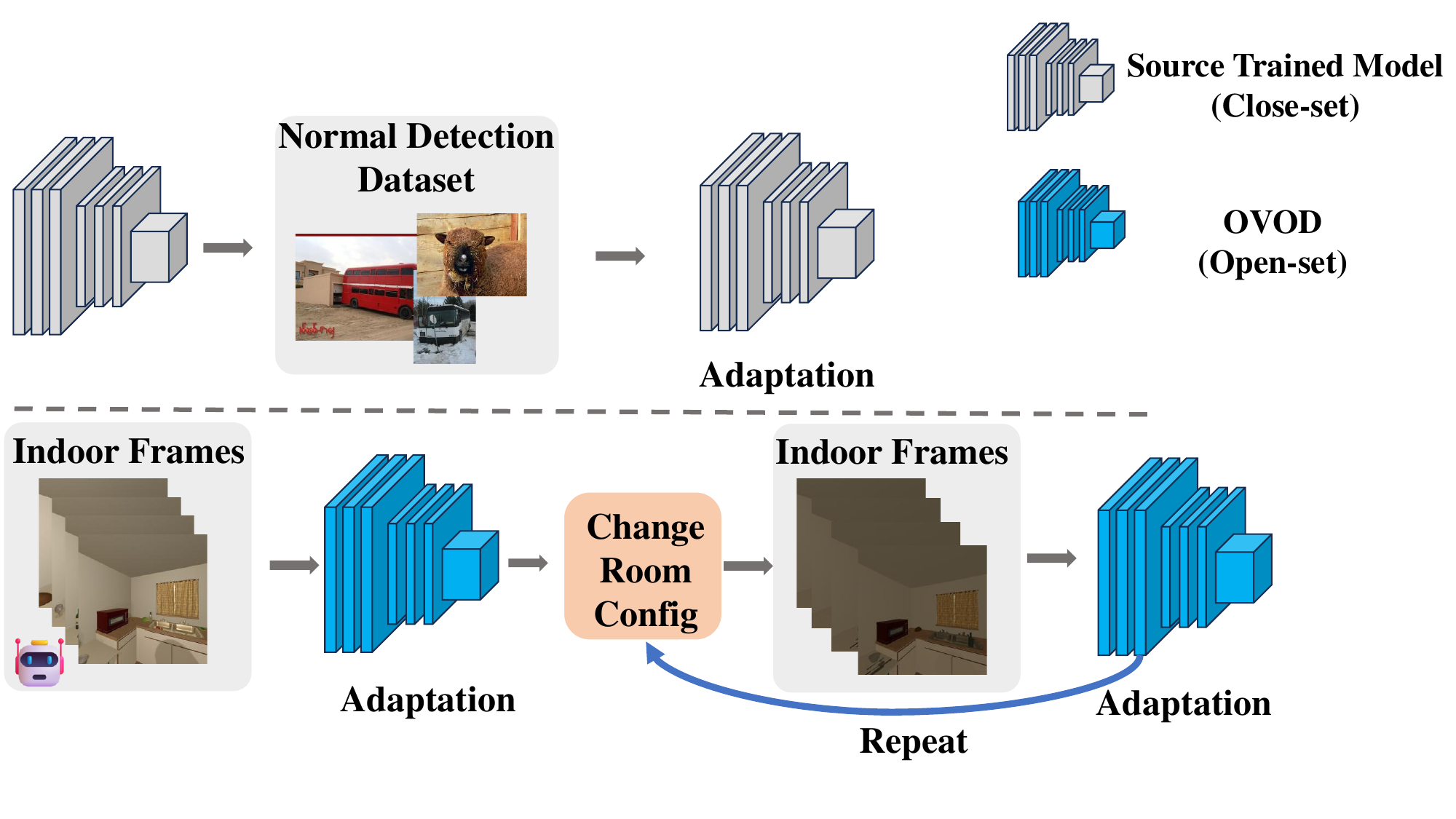}
    \vspace{0pt}
    \caption{\textbf{Top:} Traditional Source-Free Domain Adaptation (SFDA) focuses on adapting closed-form detectors to shared categories. \textbf{Bottom:} Our approach uses OVOD to adapt the model for dynamic 
 indoor environments with dynamic content and lighting conditions. The zero-shot capabilities of OVOD enable adaptation to diverse categories in these environments.}
    \label{fig:fig1}
\end{figure}

While existing SFOD methods have been primarily applied to autonomous driving datasets, they are designed for relatively structured and static environments with close-set object categories. However, these methods struggle when applied to indoor embodied perception, where the environment is highly dynamic and unstructured. Unlike outdoor scenes, indoor settings introduce distinct challenges beyond closed-set or static object configurations (Figure~\ref{fig:fig1}). \textbf{First}, the sheer diversity of real-world objects in homes and labs makes it impractical to restrict models to a predefined category set, requiring open-vocabulary capabilities.
\textbf{Second}, robotic perception in homes or labs faces continuous domain shifts as lighting, clutter, and room layouts evolve—necessitating continual adaptation rather than a one-off domain transfer. \textbf{Third}, egocentric viewpoints introduce sequential, non-stationary data, with partial occlusions and viewpoint changes demanding robust methods to exploit temporal consistency.

To address these challenges, we introduce a new benchmark called \textbf{Embodied Domain Adaptation for Object Detection (EDAOD)}. EDAOD evaluates how well Open-Vocabulary Object Detection (OVOD) systems adapt to changing room conditions—varying lighting, rearranged objects, and the sequential nature of robotic perception. This focus on incremental adaptation aligns with lifelong learning principles, where the model must continually improve as it encounters new scenarios. To enable comprehensive testing, we derive datasets from iTHOR~\cite{ithor} (simulated indoor environments) and MP3D~\cite{Matterport3D} (real-world indoor scenarios). These datasets are highly diverse in terms of object arrangements, lighting, and scene complexity, providing a robust platform to benchmark SFOD in embodied settings (see Section~\ref{sec:datasets} for details).

 Building upon EDAOD, our adaptation method tackles two primary goals: (1) refining pseudo labels through temporal instance clustering of bounding boxes across sequential frames, and (2) enhancing adaptation with a Mean Teacher framework guided by contrastive learning.

To improve label quality, we mitigate noise and mismatches using multi-scale threshold fusion, which balances high-confidence labels with broader coverage. By leveraging temporal consistency, our approach produces more accurate and stable pseudo labels, leading to improved domain adaptation. The clustered instances—appearing consistently across frames—are treated as positive samples, while unlabeled proposals serve as negatives, encouraging the model to distinguish real objects from false detections. This contrastive learning framework enhances object discrimination, resulting in more robust domain adaptation. By contrasting positive and negative samples, the model learns to refine its feature representations, improving its generalization across different indoor environments.
\noindent In this work, our main contributions are as follows:

\begin{itemize}
    \item We propose EDAOD, the first SFOD benchmark for OVOD in dynamic indoor environments.
    \item We introduce multi-scale threshold fusion for pseudo-label refinement via temporal instance clustering.
    \item We enhance adaptation using contrastive learning based on the Mean Teacher framework, yielding improved performance under non-stationary conditions.
\end{itemize}

\section{Related-work}

\subsection{Embodied Object Detection}
Embodied Object Detection aims to enhance object detection performance in physically interactive environments. 
Existing approaches can be categorized into two types: implicit memory-based embodied detection~\cite{nico} and active viewpoints optimization-based embodied detection~\cite{embod_cvpr, Fan_2024_CVPR, Fan_2024_CVPR2, bmvc_eod}. Nicolas~\cite{nico} proposed an implicit mapping technique using RGB-D data to store observed objects in the environment, which enhances the detector's performance when encountering the same objects. Klemen~\cite{embod_cvpr} explored training policies that enable embodied agents to strategically navigate environments strategically, acquiring improved viewpoints to enhance detection accuracy. Our work introduces a novel problem in embodied scenarios, emphasizing practical adaptation methods for object detectors.

\subsection{Source-free Domain Adaptation for Object Detection}

Unsupervised Domain Adaptation (UDA) adapts models trained on labelled source domains to perform effectively in unlabeled target domains~\cite{DBLP:conf/icml/LongC0J15,DBLP:journals/jmlr/GaninUAGLLML16}. When access to source data is restricted due to privacy or proprietary constraints, Source-Free Unsupervised Domain Adaptation (SFDA) enables adaptation without source data~\cite{vs2023towards,yangyang,huang2022model}. Several approaches have extended SFDA methods to object detection tasks~\cite{vs2023towards,li2021free,Shi2023ImprovingOS,hao_simplifying_2024}. For instance, Vibashan et al.~\cite{vs2023towards} proposed a method named \textbf{Memclr} that uses a cross-attention transformer-based memory module that stores the teacher model's region of interest (ROI) features from target domain samples. This module facilitates the generation of positive and negative pairs for SimCLR~\cite{simclr}-style contrastive learning, applicable in both online and offline SFDA settings for object detection. Building upon this, Vibashan~\cite{vs2022instance} introduced an Instance Relation Graph (\textbf{IRG-sfda}) that captures relationships between student and teacher-predicted proposals, which are then paired and trained using a graph contrastive loss. Shi et al.~\cite{Shi2023ImprovingOS} proposed an unsupervised data acquisition method (\textbf{UDAc-SFDA})  by selecting adapt-worth frames to enhance SFDA performance in an online manner.  Hao et al.~\cite{hao_simplifying_2024} adapted the Unbiased Teacher framework~\cite{ubteacher} for the SFDA setting, employing adaptive Batch Normalization~\cite{AdaBN} to enhance detector performance. Jiang~\cite{detic_da} was the first to propose using OVOD for domain adaptation. However, the datasets utilized were traditional object detection datasets and did not account for indoor environments.

Unlike previous SFDA methods that evaluate on shared-category datasets, our approach introduces three evaluation scenarios to assess both adaptability and generalization. Beyond the standard SFDA setting, we further examine the model’s ability to generalize across trajectories with varying lighting and object layouts, as well as its overall robustness across an entire environment (details in Section~\ref{sec:TestingScenarios}).

\subsection{Domain Adaptation for Object Detection Datasets}
Domain Adaptation for Object Detection datasets covers most of the existing detection dataset, we need to simply define what is the source domain and what is the target domain. Normally, the domain adaptation covered dataset can be divided into 4 types, General Objects datasets~\cite{coco,VOC,Clipart1k}, Self-driving datasets~\cite{cityscape,shift2022,cityscape_foggy}, Weather datasets~\cite{shift2022,cityscape_foggy}, and Face detection dataset~\cite{facedet1}. 
In contrast, we specifically target as dynamic indoor environments. These settings present unique challenges such as varying lighting conditions, occlusions, and a diverse range of object appearances and motions that are not commonly encountered in outdoor scenarios. 

\section{Preliminaries}
\subsection{Problem Formulation}
\label{sec:TestingScenarios}
In traditional Source-Free Domain Adaptation (SFDA) for object detection, a source-trained model \(\Theta_s\) is adapted to the target domain using only unlabeled target data \(\{ X_t \}\).

Our proposed approach, Embodied Domain Adaptation for Object Detection (EDAOD), leverages an open-vocabulary detector \(\Theta_o\) as the source model. Its zero-shot capabilities enable effective adaptation across diverse indoor environments. We treat each room as a distinct target domain with unique lighting, content, and object layouts. Each environment contains \(r\) layout configurations, denoted as \(\{ X_{t1}, X_{t2}, \dots, X_{tr} \}\), where \(\{ X_{ti} \}\) refers to unlabeled target frames from the \(i\)-th layout.

EDAOD introduces an iterative adaptation and evaluation process across multiple layouts:

\begin{itemize}
    \item \textbf{Same Trajectory Ability:}  
    For each layout configuration, the observed frames during exploration define the target domain \( \{ X_{ti} \} \). The model is adapted to \( \{ X_{ti} \} \) and evaluated within the same layout. This step assesses the model's ability to adapt and perform effectively in the current layout, corresponding to the traditional SFDA setting.

    \item \textbf{Next Trajectory Ability:}  
    After adaptation to a given layout configuration \( \{ X_{ti} \} \), the adapted model is directly tested on a new layout configuration \( \{ X_{ti+1} \} \) without further training. This process evaluates the model's ability to generalize across different but related configurations, progressively improving as it encounters more layouts.

    \item \textbf{Continual Adaptation and Retention:}  
    After adaptation across multiple layout configurations, the model is re-evaluated on all encountered configurations. This step checks for both continual learning and catastrophic forgetting, ensuring the model retains knowledge from previous adaptations while effectively adapting to new layouts.
\end{itemize}

This iterative adaptation and evaluation framework, EDAOD effectively addresses the domain shift challenges inherent in indoor environments. By dynamically adapting to new scenes and tasks, EDAOD enhances both cross-layout generalization and knowledge retention, making it well-suited for real-world robotic applications.

\begin{figure*}[t]
    \centering
    \includegraphics[width=\textwidth, ]{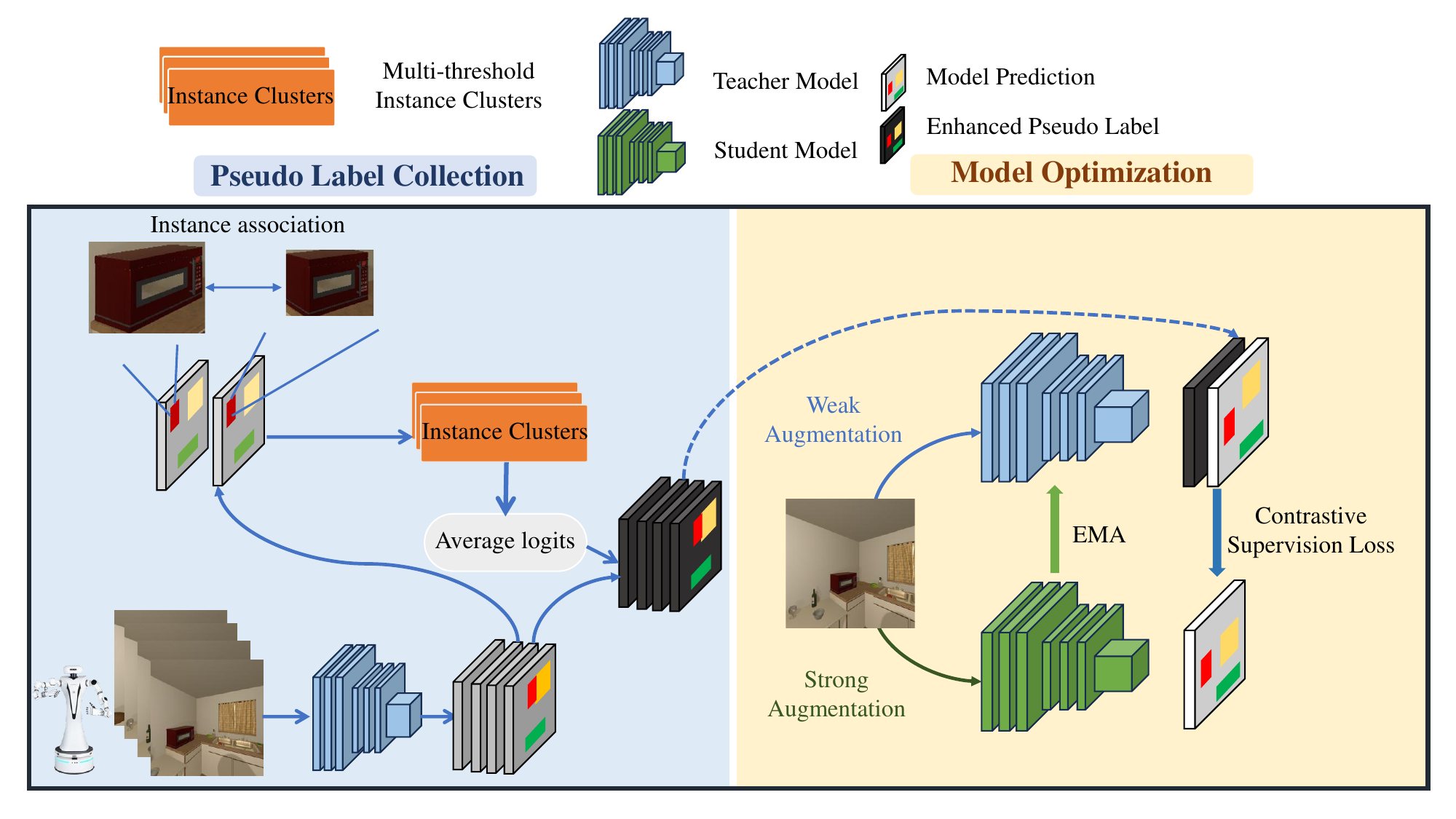}
    \vspace{0pt}
    \caption{
For each indoor scene layout, pseudo labels are first generated using the Instance Clusters algorithm to link the same instances with multiple thresholds. These pseudo labels supervise the student model in the Mean Teacher framework, ensuring domain adaptation. We further propose a contrastive loss that leverages these matched clusters to enforce feature consistency, pulling together instances of the same object while pushing apart different objects using InfoNCE-style loss.}
    \label{fig:fig2}
\end{figure*}

\section{Method}

Figure~\ref{fig:fig2} provides an overview of our proposed pipeline, where pseudo-labels are generated and used to adapt through contrastive learning to enhance domain adaptation. Our method consists of three main components: (1) a base domain adaptation model utilizing the Mean-Teacher framework~\cite{tarvainen2017mean} for open-vocabulary object detection~\cite{detic}, (2) temporal instance clustering for post-processing, and (3) a contrastive learning module to improve adaptation.

\subsection{Mean-Teacher as the Foundation for OVOD Adaptation}
We adopt the Mean Teacher framework~\cite{tarvainen2017mean} for OVOD adaptation, where both the student ($\Theta_S$) and teacher ($\Theta_T$) models are initialized from the pre-trained Detic~\cite{detic}. Given an image $I_t$, the teacher predicts weakly augmented outputs $y_T$, from which high-confidence pseudo-labels $\hat{y}_{T}$ are selected. The student model learns from these labels using strong augmentations $I_t^{s}$, optimizing the following loss:

\begin{equation}
    L_{mt} = L_{\text{rpn}} + \sum_{k=1}^{3} (L_{\text{reg}}^{(k)} + L_{\text{cls}}^{(k)})
\end{equation}

where $L_{\text{rpn}}$, $L_{\text{reg}}$, and $L_{\text{cls}}$ denote the region proposal network (RPN), regression, and classification losses across the Cascade R-CNN~\cite{cascadercnn} stages. The teacher model is updated via EMA:

\begin{equation}
    \Theta_{T,t} = \alpha_1 \Theta_{T,t-1} + (1 - \alpha_1) \Theta_{S,t-1}.
\end{equation}
The update rate is controlled by a hyperparameter \(\alpha_1\), which determines how much the teacher model adjusts after each epoch. This helps ensure the adaptation process is smooth and stable.

\subsection{Time-Consistent Instance Clustering}

Since the robot continuously explores the environment, objects detected in one frame may persist across multiple frames. To improve detection stability, we apply a temporal object-matching strategy to associate bounding boxes across consecutive frames. To leverage this temporal consistency, we associate similar objects detected across consecutive frames. Unlike other Mean Teacher-based SFDA methods~\cite{vs2023towards,Shi2023ImprovingOS} that select high-confidence predictions as pseudo-labels independently for each frame, our approach leverages temporal consistency to enhance detection stability and improve pseudo-label accuracy. During exploration, we collect the predicted bounding boxes from all frames, denoted as \( B_{1...n} \), where \( n \) represents the total number of frames. To ensure consistency in the cluster, we perform post-processing to associate bounding boxes from \( B_{1...n} \) from the pseudo-label that correspond to the same objects.

\subsubsection{Bounding Box Matching}

At the first frame, we define the bounding box set as:
\[
B_1 = \{b_1^1, b_2^1, \dots, b_{m_1}^1\}
\]
where \( m_1 \) is the total number of bounding boxes in \( B_1 \). Each bounding box \( b_i^1 \in B_1 \) is assigned to a separate cluster \( C_{1...m_1} \).

Each cluster \(C_i\) contains bounding boxes corresponding to the same object observed at different time steps, storing three key components (1) a set of matched bounding boxes, (2) their associated logit features for further cluster merging, and (3) frame id to enforce temporal consistency.

For subsequent frames, at time \( t \), we establish correspondences between the bounding boxes in the previous frame \( B_{t-1} \) and the current frame \( B_t \). This is achieved by computing the similarity between bounding boxes in both frames using:

\begin{equation}
S = \left[ s_{i,j} \right]_{m_{t-1} \times m_t} = \left[ \frac{\text{IoU}_{i,j}}{E(\mathbf{l}_{\text{prev}}[i], \mathbf{l}_{\text{curr}}[j]) + \epsilon} \right]_{m_{t-1} \times m_t}
\end{equation}

where \( S \in \mathbb{R}^{m_{t-1} \times m_t} \) represents the similarity matrix, and \( E(\mathbf{l}_{\text{prev}}[i], \mathbf{l}_{\text{curr}}[j]) \) denotes the Euclidean distance between the feature logits of bounding box \( i \) from \( B_{t-1} \) and bounding box \( j \) from \( B_t \). A small constant \( \epsilon \) is added to prevent numerical instability. 

To facilitate robust matching, we define the inverse similarity as a distance metric:

\begin{equation}
D = \left[ d_{i,j} \right]_{m_{t-1} \times m_t} =
\begin{cases} 
\frac{1}{s_{i,j}}, & \text{if } s_{i,j} \neq 0, \\
\infty, & \text{if } s_{i,j} = 0.
\end{cases}
\end{equation}

where \( D \) is the distance matrix. 

To establish correspondences between bounding boxes across frames, we construct a distance matrix \( D \) and apply the Hungarian algorithm to find the optimal assignment. If the minimum distance \( d_{i,j} \) is below the threshold \( \tau_1 \), the corresponding bounding boxes are considered the same object. If a bounding box in \( B_t \) has no valid match in \( B_{t-1} \) (distances exceed \( \tau_1 \)), it is assumed to be a newly observed object. A new cluster is then created, with this unmatched bounding box as its first element. This ensures that all detected bounding boxes are assigned to a cluster, either by extending an existing cluster or initializing a new one.

\subsubsection{Cluster Merging for Long-Term Object Matching}
In dynamic indoor environments, objects may temporarily disappear from view due to occlusions or changes in the robot's perspective. Frame-by-frame linking alone may be insufficient for robust object association. To address this, we introduce an additional cluster-merging mechanism.

After the initial bounding box matching process, we obtain multiple clusters, each containing a sequence of associated bounding boxes. To further enhance long-term object association, we merge clusters by comparing the feature similarity between their bounding boxes. Specifically, for each cluster, we compute the cosine similarity between the feature representation of the last bounding box in one cluster (\(\mathbf{f}_1\)) and the feature representation of the first bounding box in another cluster (\(\mathbf{f}_2\)), defined as $\text{sim}(\mathbf{f}_1, \mathbf{f}_2) = \frac{\mathbf{f}_1 \cdot \mathbf{f}_2}{\|\mathbf{f}_1\| \|\mathbf{f}_2\|}$.

If the similarity exceeds a predefined threshold \( \tau_2 \), the two clusters are merged, treating them as observations of the same object over time.

\subsubsection{Logit Refinement via Cluster Averaging}  
 After clustering the bounding boxes, we refine the predicted logits by computing the average logits within each cluster to improve prediction consistency and reduce fluctuations across frames. Specifically, for each cluster \( C_i \), the refined logits are given by:  

\begin{equation}
\mathbf{l}_{C_i} = \frac{1}{|C_i|} \sum_{b \in C_i} \mathbf{l}_b
\end{equation}  

where \( C_i \) denotes the \( i \)-th cluster, \( |C_i| \) is the number of bounding boxes in the cluster, and \( \mathbf{l}_{C_i} \) represents the refined logits assigned to all bounding boxes within \( C_i \). This averaging process enhances the consistency of object recognition across similar detections. 

Since every bounding box belongs to a specific cluster, we refine the pseudo-labels by averaging the logits of all bounding boxes within the same cluster. For each teacher prediction, we replace its original logit representation with the cluster-averaged logits.

\subsubsection{Multi-Scale Threshold Fusion for Robust Matching}
Selecting an appropriate threshold \( \tau_2 \) for cluster merging is challenging. A lower threshold may lead to excessive false positives due to over-linking, whereas a higher threshold reduces false positives but increases false negatives. Moreover, a fixed threshold may not generalize well across diverse environments.

To mitigate this issue, we propose a multi-scale threshold fusion approach that leverages multiple thresholds to generate a set of pseudo labels with varying levels of sensitivity: $\{\tau_{2,1}, \tau_{2,2}, \dots, \tau_{2,n}\}$

Each threshold produces a distinct set of pseudo labels, which are used to train independent models \( \{\Theta_{T,1}, \Theta_{T,2}, \dots, \Theta_{T,n}\} \). The final model \( \Theta_{Tf} \) is obtained by averaging the predictions of these models, effectively balancing precision and recall:

\begin{equation}
\Theta_{Tf} = \frac{1}{n} \sum_{i=1}^{n} \Theta_{T,i}
\end{equation}

By combining high-threshold models, which prioritize precision, with low-threshold models, which prioritize recall, this approach ensures robust performance across various environments.

\subsection{Enhancing Adaptation with Contrastive Learning}

Building on our existing framework, we incorporate contrastive learning to further improve adaptation. Given the collected cluster \( C \), where each cluster contains multiple views of the same instance, these naturally form positive pairs for contrastive learning. To leverage this structure, we introduce a contrastive learning component that utilizes bounding box features within each cluster as positive pairs. The contrastive loss is optimized using a multi-positive variant of the InfoNCE loss~\cite{InfoNCE}, enhancing the model’s ability to learn robust representations across different views.

For each set of teacher predictions \( y_{\text{T}} \) at time \( t \), we perform contrastive learning as follows:  

\subsubsection{Query (\(\mathbf{q}_i\))} 
For each bounding box \( y_{\text{T}}^i \in y_{\text{T}} \), we define the query as the feature representation extracted from the teacher model, i.e., \( \mathbf{q}_i = f_{\text{T}}(y_{\text{T}}^i) \), where \( f_{\text{T}}(\cdot) \) denotes the feature extractor of the teacher model.

\subsubsection{Positive Samples (\(\mathbf{p}_i\))} 
The positive samples consist of the features of \( y_{\text{T}}^i \) extracted by the student model and include the features of all bounding boxes within the same cluster, leading to \( \mathbf{p}_i = \{ f_{\text{S}}(y_{\text{T}}^i) \} \cup \{ f_{\text{T}}(y_{\text{T}}^j) \mid y_{\text{T}}^j \in C(y_{\text{T}}^i) \} \), where \( f_{\text{S}}(\cdot) \) represents the student model's feature extractor, and \( C(y_{\text{T}}^i) \) denotes the cluster that contains \( y_{\text{T}}^i \).

\subsubsection{Negative Samples (\( \mathbf{n}_i \))} Negative samples are drawn from bounding box proposals. Each image generates 256 proposal bounding boxes. For each proposal, features are extracted from both the teacher and student models, resulting in 512 features. To reduce computational complexity and balance the loss, we randomly select a set of 5 proposals \( \{b_k\}_{k=1}^{5} \) from the 256 proposals such that \(\text{IoU}(y_{\text{T}}^i, b_k) = 0\) for each selected proposal. The features from these proposals are used as negative samples:
$
\mathbf{n}_i = \{ f_{\text{T}}(b_k), f_{\text{S}}(b_k) \mid k = 1, \dots, 5 \}
$ where \( f_{\text{T}}(b_k) \) and \( f_{\text{S}}(b_k) \) are the features extracted from the teacher and student models, respectively.

The InfoNCE loss for each query \( y_{\text{T}}^i \) is computed as follows:  

\begin{equation}
L_{cl}^i = - \log \frac{\sum_{f \in \mathbf{p}_i} \exp(\text{sim}(\mathbf{q}_i, f) / \beta)}{\sum\limits_{f \in \mathbf{p}_i} \exp(\text{sim}(\mathbf{q}_i, f) / \beta) + \sum\limits_{f \in \mathbf{n}_i} \exp(\text{sim}(\mathbf{q}_i, f) / \beta)}
\end{equation}
where \(\text{sim}(\cdot, \cdot)\) represents the dot product, and \(\beta\) is a temperature parameter. The losses are averaged across three stages to align with the Cascade R-CNN architecture:
\begin{equation}
L_{cl} = \frac{1}{3} \sum_{k=1}^3 \left( \frac{1}{N} \sum_{i=1}^N L_{cl}^{i,k} \right)
\end{equation}
where \( k \) represents the stage index in the Cascade R-CNN model and $N$ represents the number of queries in each frame.

A Kullback–Leibler (KL) divergence loss, \( L_{kl} \), is also added to ensure alignment between the proposal classification embeddings of the student model (\( E^{\text{CLS}}_{\text{S}} \)) and those of the teacher model (\( E^{\text{CLS}}_{\text{T}} \)). To align with the Cascade R-CNN structure, this loss is formulated as: 

\begin{equation}
L_{kl} = \sum_{k=1}^3 L_{\text{KL}}(E^{\text{CLS}}_{\text{S},k}, E^{\text{CLS}}_{\text{T},k}),
\end{equation}
where \( L_{\text{KL}}(E^{\text{CLS}}_{\text{S},k}, E^{\text{CLS}}_{\text{T},k}) \) represents the Kullback–Leibler divergence between the proposal classification embeddings of the student model (\( E^{\text{CLS}}_{\text{S},k} \)) and the teacher model (\( E^{\text{CLS}}_{\text{T},k} \)) at the \( k \)-th stage.
The overall loss function is defined as:
\begin{equation}
    L_{\text{total}} = L_{mt} +  L_{cl} + L_{kl}
\end{equation}

\normalsize

\scriptsize 
\renewcommand{\arraystretch}{1.1}
\definecolor{color1}{RGB}{220, 230, 241}

\begin{table*}[]
\centering
\tiny 
\caption{AP50 results on different rooms in ITHOR under different settings}
\label{tab:ithor_results}

\resizebox{\textwidth}{!}{ 
\begin{tabular}{l|c@{\hskip 5.4pt}c@{\hskip 5.4pt}c@{\hskip 5.4pt}|c@{\hskip 5.4pt}c@{\hskip 5.4pt}c@{\hskip 5.4pt}|c@{\hskip 5.4pt}c@{\hskip 5.4pt}c@{\hskip 5.4pt}|c@{\hskip 5.4pt}c@{\hskip 5.4pt}c@{\hskip 5.4pt}|c@{\hskip 5.4pt}c@{\hskip 5.4pt}c@{\hskip 5.4pt}|c@{\hskip 5.4pt}c@{\hskip 5.4pt}c@{\hskip 5.4pt}|c@{\hskip 5.4pt}c@{\hskip 5.4pt}c@{\hskip 5.4pt}|c@{\hskip 5.4pt}c@{\hskip 5.4pt}c@{\hskip 5.4pt}}
\toprule
\multirow{2}{*}{\textbf{Method}} & \multicolumn{3}{c|}{Kitchen1} & \multicolumn{3}{c|}{Kitchen2} & \multicolumn{3}{c|}{Living Room1} & \multicolumn{3}{c|}{Living Room2} & \multicolumn{3}{c|}{Bedroom1} & \multicolumn{3}{c|}{Bedroom2} & \multicolumn{3}{c|}{Bathroom1} & \multicolumn{3}{c}{Bathroom2} \\ 
\cline{2-25}
& Same & Next & CL & Same & Next & CL & Same & Next & CL & Same & Next & CL & Same & Next & CL & Same & Next & CL & Same & Next & CL & Same & Next & CL \\ 
\hline
Source-only  & 28.42 & 28.42 & 28.42 & 33.15 & 33.15 & 33.15 & 33.88 & 33.88 & 33.88 & 37.77 & 37.77 & 37.77 & 34.27 & 34.27 & 34.27 & 38.13 & 38.13 & 38.13 & 18.12 & 18.12 & 18.12 & 30.84 & 30.84 & 30.84 \\ 
Mean-Teacher~\cite{tarvainen2017mean}  & 30.83 & 30.47 & 29.87 & 34.43 & 34.55 & 32.10 & 37.16 & 36.77 & 35.69 & 41.25 & 40.92 & 39.62 & 41.34 & 40.09 & 41.54 & 45.85 & 44.84 & 46.14 & 27.90 & 26.39 & 27.23 & 34.55 & 34.04 & 32.51 \\ 
Memclr~\cite{vs2023towards}  & 33.44 & 32.41 & 33.14 & 36.94 & 35.86 & 35.93 & 39.74 & 38.43 & 39.78 & 42.80 & 41.76 & 41.91 & 40.65 & 39.30 & 41.24 & 45.00 & 43.92 & 45.88 & 27.97 & 26.10 & 28.24 & 35.32 & 34.18 & \textbf{34.95} \\ 
Irg-sfda~\cite{vs2022instance}  & 32.16 & 31.46 & 32.50 & 36.23 & 35.22 & 35.18 & 38.53 & 37.84 & 37.66 & 42.88 & 42.17 & 42.63 & 41.25 & 39.63 & 41.88 & 45.10 & 43.99 & 46.00 & 27.46 & 25.36 & 27.69 & 34.97 & 33.97 & 34.20 \\ 
UDAc-SFDA~\cite{Shi2023ImprovingOS}  & 32.49 & 31.82 & 31.80 & 35.40 & 35.06 & 34.98 & 39.29 & 38.28 & 39.38 & 42.44 & 41.57 & 41.74 & 41.11 & 39.66 & 41.27 & 45.19 & 43.86 & 45.59 & 26.91 & 25.27 & 26.94 & 34.89 & 33.72 & 34.41 \\ 
\midrule
\rowcolor{color1}
\textbf{Ours} & \textbf{\phantom{0}36.03\phantom{0}} & \textbf{\phantom{0}34.38\phantom{0}} & \textbf{\phantom{0}35.69\phantom{0}} & \textbf{\phantom{0}37.45\phantom{0}} & \textbf{\phantom{0}36.04\phantom{0}} & \textbf{\phantom{0}37.11\phantom{0}} & \textbf{\phantom{0}41.88\phantom{0}} & \textbf{\phantom{0}39.86\phantom{0}} & \textbf{\phantom{0}40.79\phantom{0}} & \textbf{\phantom{0}45.53\phantom{0}} & \textbf{\phantom{0}44.51\phantom{0}} & \textbf{\phantom{0}45.27\phantom{0}} & \textbf{\phantom{0}42.53\phantom{0}} & \textbf{\phantom{0}40.49\phantom{0}} & \textbf{\phantom{0}42.02\phantom{0}} & \textbf{\phantom{0}47.24\phantom{0}} & \textbf{\phantom{0}45.56\phantom{0}} & \textbf{\phantom{0}47.96\phantom{0}} & \textbf{\phantom{0}31.44\phantom{0}} & \textbf{\phantom{0}28.83\phantom{0}} & \textbf{\phantom{0}33.14\phantom{0}} & \textbf{\phantom{0}36.06\phantom{0}} & \textbf{\phantom{0}35.76\phantom{0}} & 33.78 \\ 
\bottomrule
\end{tabular}
}
\end{table*}
\normalsize

\begin{table*}[t]
\centering
\caption{AP50 results on Dataset collected from MP3D under different settings}
\label{tab:mp3d_results}
\begin{tabular}{l|ccc|ccc|ccc}
\toprule
\multirow{2}{*}{\textbf{Method}} & \multicolumn{3}{c|}{Same Trajectory} & \multicolumn{3}{c|}{Next Trajectory} & \multicolumn{3}{c}{Continual Learning} \\ 
\cline{2-10}
& MP3D-A & MP3D-B & MP3D-C & MP3D-A & MP3D-B & MP3D-C & MP3D-A & MP3D-B & MP3D-C \\ 
\hline
Source-only                & 20.26  & 23.75  & 22.90  & 20.26  & 23.75  & 22.90  & 20.26  & 23.75  & 22.90  \\ 
Mean-Teacher~\cite{tarvainen2017mean}  & 18.39  & 29.84  & 33.09  & 18.73  & 29.43  & 32.74  & 17.22  & 30.16  & 32.10  \\ 
Memclr~\cite{vs2023towards}  & 18.24  & 29.87  & 33.32  & 18.50  & 29.45  & 32.61  & 17.68  & 29.56  & 33.81  \\ 
Irg-sfda~\cite{vs2022instance}  & 17.54  & 29.82  & 33.22  & 18.04  & 29.36  & 32.41  & 16.65  & 29.67  & 33.99  \\ 
UDAc-SFDA~\cite{Shi2023ImprovingOS}  & 19.27  & 29.20  & 33.98  & 19.13  & 28.89  & 32.82  & 19.03  & 29.15  & \textbf{35.16}  \\ 
\midrule
\rowcolor{color1}\textbf{Ours}  & \textbf{22.57}  & \textbf{30.68}  & \textbf{34.55}  & \textbf{21.48}  & \textbf{29.67}  & \textbf{34.11}  & \textbf{23.11}  & \textbf{31.09}  & 35.09  \\ 
\bottomrule
\end{tabular}
\end{table*}

\section{Experiment Result}

\subsection{Datasets}\label{sec:datasets}

To support our research, we utilized two datasets: {iTHOR}~\cite{ithor}, a simulated environment for embodied AI, and {Matterport3D (MP3D)}~\cite{Matterport3D}, a real-world dataset of indoor scenes. 
iTHOR is an interactive simulator designed for object manipulation tasks. It provides object detection bounding boxes for a set of pre-defined categories and allows full control over room layouts and lighting conditions, making it ideal for our experimental setup. We leveraged iTHOR to generate diverse room configurations and lighting variations manually.
However, since iTHOR consists of synthetic images, followed by~\cite{semmap}, we integrated Matterport3D (MP3D) to incorporate real-world indoor scenes and enhance dataset diversity. MP3D features real-world imagery captured from various indoor environments, complementing the simulated nature of iTHOR. 

Our study focuses solely on the target domain for both adaptation and evaluation, omitting the source domain. The dataset specifics are as follows:
\begin{itemize}
    \item \textbf{iTHOR}: The test set includes scenes from four room types: kitchens, living rooms, bedrooms, and bathrooms, each represented by two distinct environments (Kitchen1, Kitchen2, Living Room1, Living Room2, Bedroom1, Bedroom2, Bathroom1, Bathroom2). For each environment, we created five variations with different layouts and lighting conditions to simulate long-term environmental changes. Agent trajectories were manually collected using exploration policies inspired by~\cite{semmap}. While iTHOR includes 125 object categories, we excluded irrelevant ones (e.g., ``TargetCircle''), resulting in 122 categories for detection.
    \item \textbf{Matterport3D (MP3D)}: We used three distinct trajectories (MP3D-A, MP3D-B, and MP3D-C), each covering 20 object categories. Unlike iTHOR, MP3D follows the real-world domain and retains its original environments. To introduce variability, we applied five lighting conditions based on COCO-C~\cite{coco-c}.
\end{itemize}

In total, our dataset comprises 5,895 frames across 11 distinct environments, each featuring five unique layouts. Each environment is treated as an independent target domain with five unique room configurations, ensuring diversity in our domain adaptation experiments.

\subsection{Baselines}

We evaluated our method against four baselines, all of which were originally implemented for closed-set object detectors. Since their official codes are not directly applicable to our setting, we re-implemented all methods for a fair comparison. Mean-Teacher~\cite{tarvainen2017mean} serves as a foundational   SFDA in Object Detection (SFOD) baseline. MemCLR~\cite{vs2023towards} is a student-teacher SFOD framework that incorporates an external transformer block and employs contrastive learning to refine the student model. IRG-SFDA~\cite{vs2023instance} also follows a student-teacher paradigm but leverages an Instance Relation Graph to enhance adaptation. UDAc-SFDA~\cite{Shi2023ImprovingOS} is an online SFOD method that integrates a data acquisition module. Since we focus on offline setting in this work, we adapt UDAc-SFDA by removing its data acquisition process and evaluating only its proposed framework.

\begin{table*}[h]
    \centering
    \tiny 
    \caption{Ablation Study for Each Component Configuration Collected from iTHOR}
    \label{tab:abl_each}
    \renewcommand{\arraystretch}{1.2} 
    \setlength{\tabcolsep}{4pt} 
    \resizebox{\textwidth}{!}{ 
    \begin{tabular}{l|cccccccc}
        \hline
        \textbf{Method} & \textbf{Kitchen1} & \textbf{Kitchen2} & \textbf{Living Room1} & \textbf{Living Room2} & \textbf{Bedroom1} & \textbf{Bedroom2} & \textbf{Bathroom1} & \textbf{Bathroom2} \\ 
        \hline
        \rowcolor{gray!20} \multicolumn{9}{c}{\textbf{Baseline Methods}} \\
        \hline
        Source Only  & 28.42  & 33.15  & 33.88  & 37.77  & 34.27  & 38.13  & 18.12  & 30.84  \\
        Mean Teacher & 29.87  & 32.10  & 35.69  & 39.62  & 41.54  & 46.14  & 27.23  & 32.51  \\
        \hline
        \rowcolor{color1}  \multicolumn{9}{c}{\textbf{Incremental Improvements}} \\
        \hline
        Instance Cluster & 32.17  & 35.32  & 40.41  & 44.82  & 41.07  & \textbf{48.40}  & 32.15  & 32.07  \\
        Instance Cluster + Contrastive Loss & \textbf{35.69} & \textbf{37.11} & \textbf{40.79} & \textbf{45.27} & \textbf{42.02} & 47.96 & \textbf{33.14} & \textbf{33.78} \\
        \hline
    \end{tabular}}
\end{table*}

\begin{table*}[h]
    \centering
    \caption{Ablation Study for Different Hyper-Parameter Configurations}
    \label{tab:abl_hyper}
    \renewcommand{\arraystretch}{1.2} 
    \setlength{\tabcolsep}{4pt}
    \resizebox{\textwidth}{!}{ 
    \begin{tabular}{l|c|cccccccc}
        \toprule
        \textbf{Method} & \multicolumn{1}{c|}{$\tau_2$} & \textbf{Kitchen1} & \textbf{Kitchen2} & \textbf{Living Room1} & \textbf{Living Room2} & \textbf{Bedroom1} & \textbf{Bedroom2} & \textbf{Bathroom1} & \textbf{Bathroom2} \\ 
        \hline
        \multirow{4}{*}{Instance Cluster + Contrastive Loss} 
        & 0.8   & 34.58  & 36.57  & 41.74  & 44.53  & 42.40  & 46.96  & 30.22  & 35.93  \\
        & 0.85  & 34.82  & 36.87  & 41.18  & 44.79  & \textbf{42.77}  & 46.26  & 29.75  & 35.34  \\
        & 0.9   & 33.84  & 36.62  & 41.87  & 44.60  & 41.73  & 46.88  & 30.31  & 35.94  \\
        \cline{2-10}
         & \cellcolor{color1}\{0.8, 0.85, 0.9\} & \cellcolor{color1}\textbf{36.03} & \cellcolor{color1}\textbf{37.45} & \cellcolor{color1}\textbf{41.88} & \cellcolor{color1}\textbf{45.53} & \cellcolor{color1}42.53 & \cellcolor{color1}\textbf{47.24} & \cellcolor{color1}\textbf{31.44} & \cellcolor{color1}\textbf{36.06} \\
        \bottomrule
    \end{tabular}}
\end{table*}

\subsection{Implementation Details}\label{sec:Implementation_detail}
Our implementation builds upon the CenterNet2 framework~\cite{centernet2} with Detic's~\cite{detic} ResNet50~\cite{he2016deep} pre-trained backbone. Our batch size is 1 and the epoch is 15. The student model relies heavily on the teacher's predictions for adaptation. The Exponential Moving Average (EMA) parameter is set to 0.99 for \(\alpha_1\). The learning rate is set to 0.00017, and weight decay is set to 0.0001. Pseudo-labels generated by the teacher network with a confidence score above 0.5 are selected for student training. The optimization is performed using the Adam~\cite{adam} optimizer. The fuse cluster hyperparameter  $\tau_1$ is set as 1.5, $\tau_2$ we use is [0.8,0.85,0.9]. We evaluate the model performance using the mean Average Precision (mAP) calculated via the AP50 metric. The temperature $\beta$ is set as 0.1.

\subsection{Comparison with the SFDA baselines}

Table~\ref{tab:ithor_results} present the AP50 performance on the iTHOR dataset across three distinct test scenarios: testing on the same trajectory, testing on the next trajectory, and continual learning. Each room in the tables was evaluated independently, with the source model used for initialization. Our model demonstrates superior performance compared to the baseline model in most scenarios across various environments. Notably, our model outperforms the baseline model in all environments when tested on both the same and next trajectories. This result highlights the effectiveness of our approach in adapting to the current trajectory and seamlessly transitioning to perform well on the subsequent trajectory without additional adaptation. 

In Table~\ref{tab:mp3d_results}, our method is the only one that outperforms the Source-only baseline in MP3D-A. Moreover, it demonstrates improvements in most scenarios, including the same trajectory, next trajectory, and continual learning (CL). In the CL scenario, the model adapts to all trajectories within the current scene and is then evaluated across all paths to assess whether it retains knowledge of the initial trajectories. This result highlights the robustness of our approach in avoiding catastrophic forgetting while achieving consistent gains.

\begin{figure}[h]
    \centering
    \includegraphics[width=0.5\textwidth, height=0.30\textwidth]{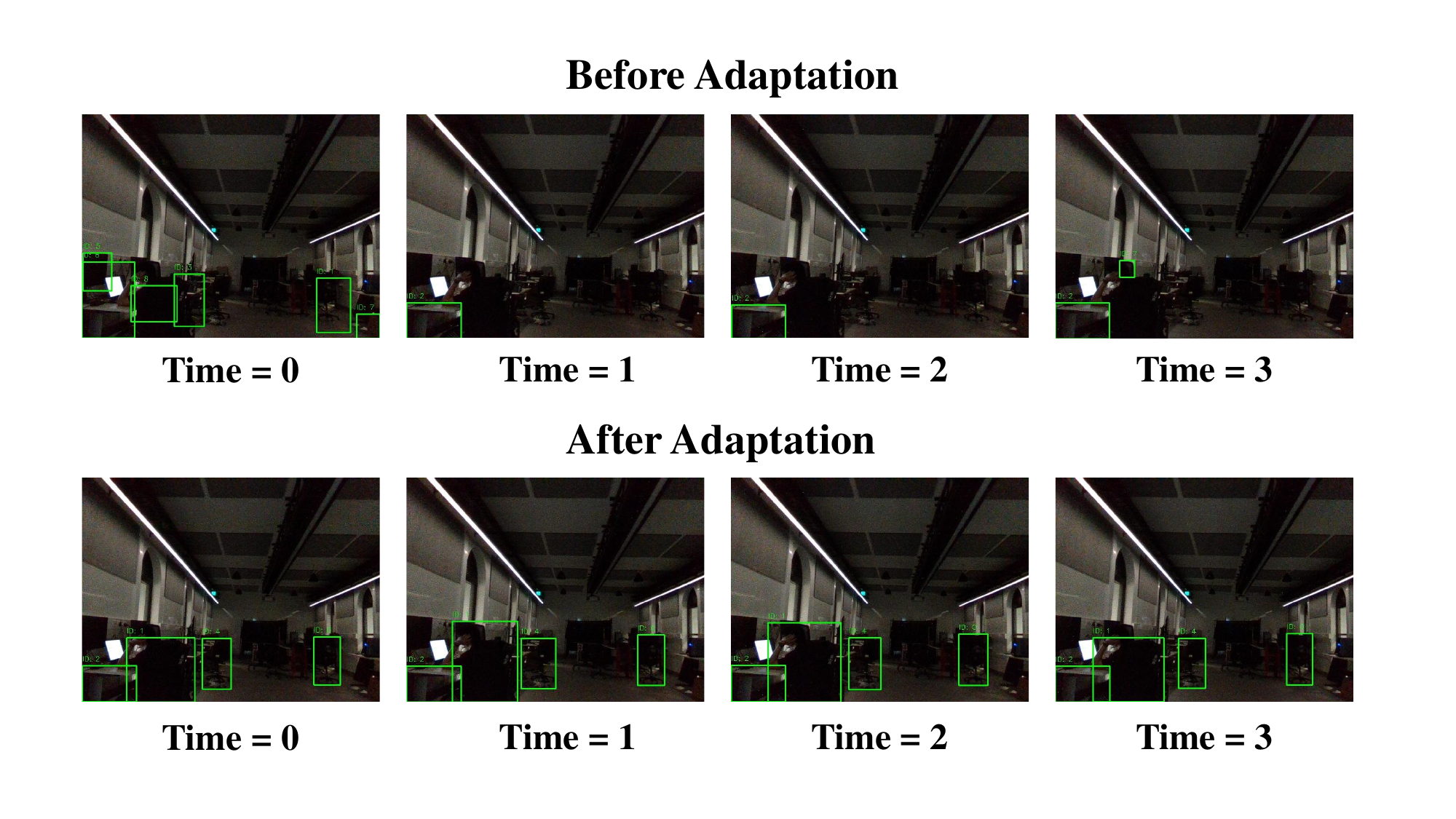}
    \vspace{-5mm}

    \caption{The top row shows the predictions for the first four frames in our lab under the lowest-light condition environment. The bottom row presents the results after adaptation, where the detector demonstrates improved stability and detects more objects accurately.}
    \label{fig:fig3}
\end{figure}
\subsection{Ablation Study}
In Table~\ref{tab:abl_each}, we compare the performance of our model under different configurations using the CL setting. ``Instance Cluster'' refers to optimizing the model with improved pseudo labels, while ``Instance Cluster + Contrastive Loss'' incorporates our proposed effective adaptation method with improved pseudo-label. Our model achieves scores that surpass both Mean-Teacher and Source-Only baselines when using only ``Instance Cluster''. Furthermore, when we incorporate our proposed adaptation method, the performance improves even further, demonstrating the effectiveness of our approach.

We evaluate the importance of the fuse cluster for the Same Trajectory. Specifically, we fuse $\tau_2$ values of [0.8, 0.85, 0.9] as our final result. As shown in Table~\ref{tab:abl_hyper}, using any single parameter results in worse performance compared to the fused parameters. This clearly demonstrates that incorporating multi-scale threshold fusion enables our model to balance the trade-offs between thresholds: lower thresholds result in longer but noisier clusters, while higher thresholds reduce noise but are prone to premature interruptions. Consequently, our approach outperforms single-scale configurations.
\begin{table}[t]
    \centering
    \caption{Detection Performance Before and After Adaptation in Real-World Environments (Captured Under lowest-light condition Environments)}
    \label{tab:real_robot_average}
    \renewcommand{\arraystretch}{1.2} 
    \setlength{\tabcolsep}{5pt} 
    \resizebox{0.4\textwidth}{!}{ 
    \begin{tabular}{l|cc|c}
        \toprule
        \multirow{2}{*}{\textbf{Object Category}} & \multicolumn{2}{c|}{\textbf{Adaptation (\#Detections)}} & \multirow{2}{*}{\textbf{\#Ground Truth}} \\ 
        \cline{2-3}
        & \textbf{Before} & \textbf{After} &  \\ 
        \hline
        Chair   & 54  & \textbf{117}  & 120  \\
        Table   & 37  & \textbf{51}   & 220  \\
        Toolbox & 16  & \textbf{22}   & 100  \\
        \hline
        \textbf{Average} & 35.67  & \textbf{63.33} & 146.67  \\
        \bottomrule
    \end{tabular}}
\end{table}

\subsection{Real World Robot Visualization}

We evaluate our model in a real-world environment within our laboratory by deploying the detection model on mobile robot equipped with an Intel RealSense D435 camera. Unlike the simulator, where frames are collected sequentially at consistent distances and angles, real-world data collection is influenced by variations in the robot’s movement speed and potential wireless communication delays, leading to slight inconsistencies in frame intervals. 

Our trajectory data is collected across three different environmental settings, each with variations in lighting and object layouts. To assess the adaptability of our model in challenging conditions, we specifically evaluate it using images captured under the lowest-light condition. This setting allows us to analyze how well the adapted model performs in extreme environments, where some objects may be less frequently observed due to lighting limitations.

The results in Table~\ref{tab:real_robot_average} demonstrate that adaptation significantly improves detection recall. The number of detected objects increased across all scenarios, with an average improvement from 35.67 to 63.33 detections. This finding demonstrates that adaptation significantly improves detection accuracy, allowing the model to better handle challenging environments and increasing its reliability for real-world deployment.

The visualization of the lowest-light condition environment is shown in Figure~\ref{fig:fig3}. It clearly demonstrates that after adaptation, the detection performance improves, indicating that the detector becomes more familiar with the current environment and adapts more effectively to challenging lighting conditions.

\section{CONCLUSION}

In this work, we proposed Embodied Domain Adaptation for Object Detection (EDAOD), a novel Source-Free Domain Adaptation (SFDA) approach tailored for indoor robotic perception. To establish a benchmark, we collected an embodied dataset by modifying room configurations and introducing diverse domain shifts. Our method enhances object detection adaptation through temporal clustering, multi-scale threshold fusion, and contrastive learning built upon the Mean Teacher framework.

Extensive experiments demonstrate that EDAOD outperforms existing SFDA methods in adapting to dynamic indoor environments, achieving significant performance gains without requiring access to source data. The proposed approach helps alleviate domain shifts and improves adaptation in indoor robotic perception, demonstrating its potential for real-world applications.

In future work, we aim to explore adaptive strategies for long-term robot learning and investigate the integration of active perception techniques to enhance adaptation in dynamic environments.
\bibliographystyle{IEEEtran}
\bibliography{ref.bib}

\begin{thebibliography}{10}
\providecommand{\url}[1]{#1}
\csname url@samestyle\endcsname
\providecommand{\newblock}{\relax}
\providecommand{\bibinfo}[2]{#2}
\providecommand{\BIBentrySTDinterwordspacing}{\spaceskip=0pt\relax}
\providecommand{\BIBentryALTinterwordstretchfactor}{4}
\providecommand{\BIBentryALTinterwordspacing}{\spaceskip=\fontdimen2\font plus
\BIBentryALTinterwordstretchfactor\fontdimen3\font minus \fontdimen4\font\relax}
\providecommand{\BIBforeignlanguage}[2]{{%
\expandafter\ifx\csname l@#1\endcsname\relax
\typeout{** WARNING: IEEEtran.bst: No hyphenation pattern has been}%
\typeout{** loaded for the language `#1'. Using the pattern for}%
\typeout{** the default language instead.}%
\else
\language=\csname l@#1\endcsname
\fi
#2}}
\providecommand{\BIBdecl}{\relax}
\BIBdecl

\bibitem{coco}
T.~Lin \emph{et~al.}, ``Microsoft {COCO:} common objects in context,'' in \emph{ECCV}, vol. 8693.\hskip 1em plus 0.5em minus 0.4em\relax Springer, 2014, pp. 740--755.

\bibitem{detic}
X.~Zhou \emph{et~al.}, ``Detecting twenty-thousand classes using image-level supervision,'' in \emph{ECCV}, 2022.

\bibitem{yoloworld}
T.~Cheng \emph{et~al.}, ``Yolo-world: Real-time open-vocabulary object detection,'' in \emph{CVPR}, 2024.

\bibitem{gdino}
S.~Liu \emph{et~al.}, ``Grounding {DINO:} marrying {DINO} with grounded pre-training for open-set object detection,'' \emph{CoRR}, vol. abs/2303.05499, 2023.

\bibitem{chen2018domain}
Y.~Chen \emph{et~al.}, ``Domain adaptive faster r-cnn for object detection in the wild,'' in \emph{CVPR}, 2018, pp. 3339--3348.

\bibitem{saito2019strongweak}
K.~Saito \emph{et~al.}, ``Strong-weak distribution alignment for adaptive object detection,'' in \emph{CVPR}, 2019, pp. 6956--6965.

\bibitem{sindagi2020prior}
V.~A. Sindagi \emph{et~al.}, ``Prior-based domain adaptive object detection for hazy and rainy conditions,'' in \emph{ECCV}.\hskip 1em plus 0.5em minus 0.4em\relax Springer, 2020, pp. 763--780.

\bibitem{udaa}
G.~Mattolin \emph{et~al.}, ``Confmix: Unsupervised domain adaptation for object detection via confidence-based mixing,'' in \emph{WACV}.\hskip 1em plus 0.5em minus 0.4em\relax {IEEE}, 2023, pp. 423--433.

\bibitem{vs2023towards}
V.~VS \emph{et~al.}, ``Towards online domain adaptive object detection,'' in \emph{CVPR}, 2023, pp. 478--488.

\bibitem{tarvainen2017mean}
A.~Tarvainen \emph{et~al.}, ``Mean teachers are better role models: Weight-averaged consistency targets improve semi-supervised deep learning results,'' \emph{NeurIPS}, vol.~30, 2017.

\bibitem{yangyang}
Y.~Shu \emph{et~al.}, ``Source-free unsupervised domain adaptation with hypothesis consolidation of prediction rationale,'' \emph{CoRR}, 2024.

\bibitem{cityscape}
M.~Cordts \emph{et~al.}, ``The cityscapes dataset for semantic urban scene understanding,'' in \emph{CVPR}, 2016, pp. 3213--3223.

\bibitem{sim10k}
M.~Johnson-Roberson \emph{et~al.}, ``Driving in the matrix: Can virtual worlds replace human-generated annotations for real world tasks?'' \emph{arXiv preprint arXiv:1610.01983}, 2016.

\bibitem{shift2022}
T.~Sun \emph{et~al.}, ``{SHIFT:} a synthetic driving dataset for continuous multi-task domain adaptation,'' in \emph{CVPR}, June 2022, pp. 21\,371--21\,382.

\bibitem{VOC}
M.~Everingham \emph{et~al.}, ``The pascal visual object classes {(VOC)} challenge,'' \emph{IJCV}, 2010.

\bibitem{ithor}
E.~Kolve \emph{et~al.}, ``{AI2-THOR:} an interactive 3d environment for visual {AI},'' \emph{CoRR}, vol. abs/1712.05474, 2017.

\bibitem{Matterport3D}
A.~Chang \emph{et~al.}, ``Matterport3d: Learning from rgb-d data in indoor environments,'' \emph{3DV}, 2017.

\bibitem{nico}
N.~H. Chapman \emph{et~al.}, ``Enhancing embodied object detection through language-image pre-training and implicit object memory,'' \emph{CoRR}, vol. abs/2402.03721, 2024.

\bibitem{embod_cvpr}
K.~Kotar \emph{et~al.}, ``Interactron: Embodied adaptive object detection,'' in \emph{CVPR}.\hskip 1em plus 0.5em minus 0.4em\relax {IEEE}, 2022, pp. 14\,840--14\,849.

\bibitem{Fan_2024_CVPR}
L.~Fan \emph{et~al.}, ``Evidential active recognition: Intelligent and prudent open-world embodied perception,'' in \emph{CVPR}, June 2024, pp. 16\,351--16\,361.

\bibitem{Fan_2024_CVPR2}
L.~Fan and et~al., ``Active open-vocabulary recognition: Let intelligent moving mitigate clip limitations,'' \emph{CVPR}, pp. 16\,394--16\,403, June 2024.

\bibitem{bmvc_eod}
Z.~Fang \emph{et~al.}, ``Move to see better: Self-improving embodied object detection,'' in \emph{BMVC}.\hskip 1em plus 0.5em minus 0.4em\relax {BMVA} Press, 2021, p. 270.

\bibitem{DBLP:conf/icml/LongC0J15}
M.~Long \emph{et~al.}, ``Learning transferable features with deep adaptation networks,'' in \emph{ICML}, 2015.

\bibitem{DBLP:journals/jmlr/GaninUAGLLML16}
Y.~Ganin \emph{et~al.}, ``Domain-adversarial training of neural networks,'' \emph{JMLR}, vol.~17, pp. 59:1--59:35, 2016.

\bibitem{huang2022model}
J.~Huang \emph{et~al.}, ``Model adaptation: Historical contrastive learning for unsupervised domain adaptation without source data,'' in \emph{NeurIPS}, 2021, pp. 3635--3649.

\bibitem{li2021free}
X.~Li \emph{et~al.}, ``A free lunch for unsupervised domain adaptive object detection without source data,'' in \emph{AAAI}, vol.~35, no.~10, 2021, pp. 8474--8481.

\bibitem{Shi2023ImprovingOS}
X.~Shi \emph{et~al.}, ``Improving online source-free domain adaptation for object detection by unsupervised data acquisition,'' in \emph{ECCVW}, 2024, p. abs/2310.19258.

\bibitem{hao_simplifying_2024}
Y.~Hao \emph{et~al.}, ``Simplifying source-free domain adaptation for object detection: Effective self-training strategies and performance insights,'' in \emph{ECCV 2024}, 2024.

\bibitem{simclr}
T.~Chen \emph{et~al.}, ``A simple framework for contrastive learning of visual representations,'' in \emph{ICML}, ser. Proceedings of Machine Learning Research, 2020.

\bibitem{vs2022instance}
V.~VS \emph{et~al.}, ``Instance relation graph guided source-free domain adaptive object detection,'' \emph{CVPR}, 2023.

\bibitem{ubteacher}
Y.~Liu \emph{et~al.}, ``Unbiased teacher for semi-supervised object detection,'' in \emph{ICLR}, 2021.

\bibitem{AdaBN}
Y.~Li \emph{et~al.}, ``Adaptive batch normalization for practical domain adaptation,'' \emph{PR}, vol.~80, pp. 109--117, 2018.

\bibitem{detic_da}
K.~Jiang \emph{et~al.}, ``Domain adaptation for large-vocabulary object detectors,'' \emph{CoRR}, vol. abs/2401.06969, 2024.

\bibitem{Clipart1k}
N.~Inoue \emph{et~al.}, ``Cross-domain weakly-supervised object detection through progressive domain adaptation,'' in \emph{CVPR}, 2018, pp. 5001--5009.

\bibitem{cityscape_foggy}
C.~Sakaridis \emph{et~al.}, ``Semantic foggy scene understanding with synthetic data,'' \emph{IJCV}, vol. 126, pp. 973--992, 2018.

\bibitem{facedet1}
H.~Nada \emph{et~al.}, ``Pushing the limits of unconstrained face detection: a challenge dataset and baseline results,'' in \emph{BTAS}.\hskip 1em plus 0.5em minus 0.4em\relax {IEEE}, 2018, pp. 1--10.

\bibitem{cascadercnn}
Z.~Cai \emph{et~al.}, ``Cascade {R-CNN:} delving into high quality object detection,'' in \emph{CVPR}, 2018, pp. 6154--6162.

\bibitem{InfoNCE}
A.~van~den Oord \emph{et~al.}, ``Representation learning with contrastive predictive coding,'' \emph{CoRR}, vol. abs/1807.03748, 2018.

\bibitem{semmap}
V.~Cartillier \emph{et~al.}, ``Semantic mapnet: Building allocentric semantic maps and representations from egocentric views,'' in \emph{AAAI}, 2021, pp. 964--972.

\bibitem{coco-c}
C.~Michaelis \emph{et~al.}, ``Benchmarking robustness in object detection: Autonomous driving when winter is coming,'' \emph{arXiv preprint arXiv:1907.07484}, 2019.

\bibitem{vs2023instance}
V.~VS \emph{et~al.}, ``Instance relation graph guided source-free domain adaptive object detection,'' in \emph{Proceedings of the IEEE/CVF Conference on Computer Vision and Pattern Recognition}, 2023, pp. 3520--3530.

\bibitem{centernet2}
X.~Zhou \emph{et~al.}, ``Probabilistic two-stage detection,'' \emph{CoRR}, vol. abs/2103.07461, 2021.

\bibitem{he2016deep}
K.~He \emph{et~al.}, ``Deep residual learning for image recognition,'' in \emph{CVPR}, 2016, pp. 770--778.

\bibitem{adam}
D.~P. Kingma \emph{et~al.}, ``Adam: {A} method for stochastic optimization,'' in \emph{ICLR}, 2015.

\end{thebibliography}
\end{document}